\documentclass[11pt,a4paper]{article}
\usepackage{times,latexsym}
\usepackage{url}
\usepackage[T1]{fontenc}
 \usepackage[dvipsnames]{xcolor}
 \usepackage{inconsolata}

\usepackage[acceptedWithA]{tacl2021v1}

\usepackage{xspace,mfirstuc,tabulary}

\newif\iftaclinstructions
\taclinstructionsfalse % AUTHORS: do NOT set this to true
\iftaclinstructions
\renewcommand{\confidential}{}
\renewcommand{\anonsubtext}{(No author info supplied here, for consistency with
TACL-submission anonymization requirements)}
\newcommand{\instr}
\fi

\iftaclpubformat % this "if" is set by the choice of options

\else

\fi

\usepackage{times}
\usepackage{latexsym}
\usepackage{amsmath}
\usepackage{amsfonts}
\usepackage{amsthm}
\usepackage{url}
\usepackage{tikz}
\usepackage{xspace}
\usetikzlibrary{bayesnet}
\usepackage{adjustbox}
\usepackage{booktabs}
\usepackage{mathtools}
\usepackage{xfrac}
\usepackage{comment}
\usepackage{todonotes}
\usepackage{thmtools} 
\usepackage{thm-restate}
\usepackage[shortlabels]{enumitem}
\usepackage{linguex}
\usepackage{bbm}

\usepackage{cleveref}
\crefname{section}{\S}{\S\S}
\crefname{table}{Table}{}
\crefname{figure}{Figure}{}
\crefname{algorithm}{Alg.}{}
\crefname{equation}{Eq.}{Eq.}
\crefname{appendix}{App.}{}
\crefname{theorem}{Theorem}{}
\crefname{prop}{Proposition}{}
\crefname{cor}{Corollary}{}
\crefname{observation}{Observation}{}
\crefname{assumption}{Assumption}{}
\crefname{hypothesis}{Hyp.}{Hypotheses}
\crefformat{section}{\S#2#1#3}

\makeatletter
\newcommand*\iftodonotes{\if@todonotes@disabled\expandafter\@secondoftwo\else\expandafter\@firstoftwo\fi}  % defines \iftodonotes{<true>}{<false>}, thanks to https://tex.stackexchange.com/questions/126559/conditional-based-on-packageoption
\makeatother

\newenvironment{sproof}{\noindent{\\\textit{Proof}:}}{\hfill$\blacksquare$\\}
\newenvironment{mproof}{\noindent{\textit{Proof}:}}{\hfill$\blacksquare$}

\newcommand{\vecw}{{\boldsymbol w}}

\newcommand{\defeq}{\overset{\text{\tiny def}}{=}}
\newcommand{\MI}{\mathrm{MI}}
\newcommand{\MIdo}{\mathrm{MI}_{\mathrm{do}}}

\newcommand{\nouns}{{\color{MyBlue}\boldsymbol{\mathcal{N}}}}
\newcommand{\adjs}{{\color{MyPurple}\mathcal{A}}}
\newcommand{\adj}{{\color{MyPurple}a}}
\newcommand{\adjtwo}{{\color{MyPurple}b}}

\newcommand{\genmsc}{{\color{OliveGreen} \textsc{msc}}}
\newcommand{\genfem}{{\color{OliveGreen} \textsc{fem}}}
\newcommand{\genneu}{{\color{OliveGreen} \textsc{neu}}}
\newcommand{\mathcomment}[1]{{\color{gray} #1}}
\newcommand{\vtheta}{\boldsymbol{\theta}}

\newcommand{\nounseval}{\widetilde{\setN}}

\newcommand{\embed}{\mathbf{e}}

\newcommand{\defn}[1]{{\textbf{#1}}}

\newcommand{\gender}{{\color{OliveGreen} g}}
\newcommand{\gendertwo}{{\color{OliveGreen} h}}
\newcommand{\genders}{{\color{OliveGreen}\mathcal{G}}}
\newcommand{\rv}[1]{{#1}}
\newcommand{\setfont}[1]{\mathrm{#1}}
\newcommand{\rvG}{{\color{OliveGreen}\rv{G}}}
\newcommand{\rvA}{{\color{MyPurple}\rv{A}}}
\newcommand{\rvN}{{\color{MyBlue}\rv{N}}}

\newcommand{\setA}{{\color{MyPurple}\setfont{A}}}
\newcommand{\setN}{{\color{MyBlue}\setfont{N}}}

\newcommand{\R}{\mathbb{R}}
\newcommand{\noun}{{\color{MyBlue}\boldsymbol{n}}}
\newcommand{\nountwo}{{\color{MyBlue}\boldsymbol{m}}}

\definecolor{MyPurple}{RGB}{149,43,96}
\definecolor{MyBlue}{RGB}{37,111,174}
\definecolor{OliveGreen}{RGB}{117,157,139}
\definecolor{MyOrange}{RGB}{217,121,44}
\newcommand{\word}[1]{{\color{MyOrange} \textit{#1}}}

\newcommand{\XX}{{five}\xspace}

\newcommand{\JSD}{\mathrm{JS}}
\newcommand{\KL}{\mathrm{KL}}

\newcommand{\Hdo}{\mathrm{H}_{\mathrm{do}}}

\newcommand{\pgender}{p_{\rvG}}
\newcommand{\pnoun}{p_{\rvN}}
\newcommand{\padj}{p_{\rvA}}

\newcommand{\Dtrn}{\mathrm{D}_{\text{trn}}}
\newcommand{\Dtst}{\mathrm{D}_{\text{tst}}}

\newcommand{\spacebefore}{\,\,\,\,\,\,\,\,\,}

\newcommand{\expnumber}[2]{{#1}\mathrm{e}{#2}}

\newcommand{\citeposs}[1]{\citeauthor{#1}'s (\citeyear{#1})}

\usepackage{amssymb}

\title{The Causal Influence of Grammatical Gender on Distributional Semantics}
\author{
Karolina Stańczak$^{1,2,3,5}$\quad
Kevin Du$^{3}$ \quad \\
\textbf{Adina Williams}$^{4}$ \quad
\textbf{Isabelle Augenstein}$^{5}$ \quad
\textbf{Ryan Cotterell}$^{3}$ \\
\small{$^1$Mila – Quebec AI Institute\quad
$^2$McGill University\quad
$^3$ETH Zurich\quad
$^4$FAIR at Meta AI\quad
$^5$University of Copenhagen}\\
\small{{\tt \href{mailto:karolina.stanczak@mila.quebec}{karolina.stanczak@mila.quebec}} \quad
{\tt \href{mailto:kevidu@ethz.ch}{kevin.du@inf.ethz.ch}}} \quad \\
\small{\tt \href{mailto:adinawilliams@meta.com}{adinawilliams@meta.com}} \quad 
{\tt \href{mailto:augenstein@di.ku.dk}{augenstein@di.ku.dk}} \quad 
{\tt \href{mailto:rcotterell@inf.ethz.ch}{rcotterell@inf.ethz.ch}} \quad 
}

\begin{document}
\maketitle
% % Neo-whorfian centric
% \begin{abstract}
% The relation between the grammatical gender of nouns and the concepts those nouns represent is a controversial subject in modern linguistics. On one side, researchers claim that grammatical gender \emph{does}
% influence how we describe nouns, including inanimate ones which we call the neo-Whorfian hypothesis. 
% On the other side, many linguists resist this idea and argue that how humans talk about nouns is independent of their inherent properties, such as gender. Experimentally, most evidence adduced in support of the neo-Whorfian position has come from small-scale controlled laboratory experiments. In this work, we offer a new data-driven experiment with machine learning techniques from causal inference to shed more light on this question. We find that...

% % We test whether the grammatical genders assigned to nouns have semantic consequences. 
% % People's ideas about the genders of objects are strongly influenced by the grammatical genders assigned to these objects in their native language
% % the influence on the way people think about inanimate objects
% \end{abstract}

\begin{abstract}
% 1) Gender in some languages is related to lexical semantics while in others it's not
% 2) Neo-Whorfianism and conflicting studies
% 3) Prior studies might have overestimated the effect of gender because they didn't control for the noun meaning
% 4) Our causal method
% 5) Results

How much meaning influences gender assignment across languages is an active area of research in linguistics and cognitive science.
We can view current approaches as aiming to determine where gender assignment falls on a spectrum, from being fully arbitrarily determined to being largely semantically determined. 
For the latter case, there is a formulation of the neo-Whorfian hypothesis, which claims that even inanimate noun gender influences how people conceive of and talk about objects (using the choice of adjective used to modify inanimate nouns as a proxy for meaning).
We offer a novel, causal graphical model that jointly represents the interactions between a noun's grammatical gender, its meaning, and adjective choice. 
In accordance with past results, we find a significant relationship between the gender of nouns and the adjectives that modify them.
However, when we control for the meaning of the noun, the relationship between grammatical gender and adjective choice is near zero and insignificant.\looseness=-1

% relevant to both neo-Whorfianism and the grammatical gender assignment problem, namely gender, adjectives, and noun meaning

% We test whether the grammatical genders assigned to nouns have semantic consequences. 
% People's ideas about the genders of objects are strongly influenced by the grammatical genders assigned to these objects in their native language
% the influence on the way people think about inanimate objects
\end{abstract}

\section{Introduction}
\label{sec:introduction}
Approximately half of the world's languages have grammatical gender \citep{wals-30}, a grammatical phenomenon that groups nouns together into classes that share morphosyntactic properties \citep{hockett-1958-course, corbett1991gender, kramer-2015-morphosyntax}. 
Among languages that have gender, there is variation in the number of gender classes; for example, some languages have only two classes, e.g., all Danish nouns are classed as either common or neuter, whereas others have significantly more, e.g., Nigerian Fula has around 20, depending on the variety \citep{Arnott-1967, Koval-1979, Breedveld-1995}.
Languages also vary with respect to how much gender assignment, i.e., how nouns are sorted into particular genders, is related to the form and the meaning of the noun \citep{corbett1991gender, plaster-polinky-2007-women, wals-32, corbett-2014-gender, kramer-2020-grammatical, sahai-sharma-2021-predicting}.
Some languages group nouns into gender classes that are highly predictable from phonological \citep{Parker-and-Hayward-1985, corbett1991gender, wals-32} or morphological  \citep{corbett1991gender,wals-32, corbett-fraser-2000-systems} information, while others, such as the Dagestanian languages Godoberi and Bagwalal, seem to be predictable from meaning \citep{corbett1991gender,corbett-fraser-2000-systems, corbett-2014-gender}---although, even for most of the strictly semantic systems, there are exceptions.\looseness=-1

Despite this variation, gender assignment is rarely, if ever, wholly predictable from meaning alone. 
In many languages, there is a semantic core of nouns that are conceptually coherent \citep{aksenov-1984, corbett1991gender, williams-etal-2019-quantifying, kramer-2020-grammatical} and a surround that is somewhat less semantically coherent. 
Axes along which genders are conceptually coherent often include semantic properties of animate nouns, with inanimate nouns appearing in the surround. 
For example, in Spanish, despite the fact that the nouns \word{table} (\word{mesa} in Spanish) and \word{woman} (\word{mujer} in Spanish) appear in the same gender (i.e., feminine), it is hard to imagine what meaning they share. 
Indeed, some linguists posit that gender assignment for inanimate nouns is effectively arbitrary \citep{bloomfield1935language,YAikhenvald2000-YAICAT-2,foundalis-2002-evolution}.
And, to the extent that gender assignment is \emph{not} fully arbitrary for inanimate nouns \citep{williams-etal-2021-relationships}, many researchers argue there is no compelling evidence showing grammatical gender affects how we conceptualize objects \citep{samuel-2019-grammatical} or the distributional properties of language \citep{mickan2014key}.\looseness=-1

%In short, under this kind of view, which we call the `arbitrariness of gender' view, expects no transparent relationship between grammatical gender and how people use inanimate nouns.\looseness=-1
%Such arbitrariness leads some linguists to conclude that the grammatical properties of nouns, such as gender, are probably independent of how we talk and think about the objects nouns refer to \citep{sapir1921language}.

However, not all researchers agree that non-arbitrariness in gender assignment, to the extent it exists, should be assumed to have no bearing on language production. 
\citet{boroditsky2003linguistic} famously argued for a \emph{causal} relationship between the gender assigned to inanimate nouns and their usage, in a view colloquially known as the neo-Whorfian hypothesis after Benjamin Whorf \citep{Whorf1956language}.
Proponents of this view have studied human associations, under the assumption that people's perceptions of the genders of objects are strongly influenced by the grammatical genders these objects are assigned in their native language \citep{boroditsky2003sex,Semenuks2017EffectsOG}. 
%Proponents of this view have focused on adjectives as their dependent variable, under the assumption that the gender of inanimate nouns may influence how adjectives that modify them are selected \citep{boroditsky2003sex,Semenuks2017EffectsOG}. 
One manifestation of this perception is the choice of adjectives used to describe nouns \citep{Semenuks2017EffectsOG}.
While this is an intriguing possibility, there are additional lexical properties of nouns that may act as confounders and, thus, finding statistical evidence for the causal effect of grammatical gender on adjective choice requires great care.\looseness=-1
%Until we can prevent noun meaning from confounding our experiments, we will be unable to argue definitively for or against the causal neo-Whorfian view.\looseness=-1

% Our solution/method
To facilitate a cleaner way to reason about the causal influence grammatical gender may have on adjective usage, we introduce a causal graphical model to represent the interactions between an inanimate noun's grammatical gender, its meaning, and the choice of its descriptors. 
This causal framework enables intervening on the values of specific factors to isolate the effects between various properties of languages. 
Our model explains the distribution of adjectives that modify a noun, conditioned on both a representation of the noun's meaning and the gender of the noun itself. 
Upon estimation of the parameters of the causal graphical model, we test the neo-Whorfian hypothesis beyond the anecdotal level.
First, we validate our model by comparing it to the method presented in prior work without any causal intervention. 
%examine whether the assignment of grammatical gender is related to the noun's meaning, i.e., whether it is, in fact, non-arbitrary.
Second, we employ our model with a causal intervention on the noun meaning to test the neo-Whorfian hypothesis.
That is, we ask a counterfactual question:
Had nouns been lexicalized with different grammatical genders but retained their same meanings, would the distribution of adjectives that speakers use to modify them have been different? \looseness=-1
%We quantify this difference in distributions information-theoretically, using the Jensen--Shannon divergence.\looseness=-1

% results
We employ our model on \XX languages that exhibit grammatical gender: four Indo-European languages (German, Polish, Portuguese, and Spanish) and one language from the Afro-Asiatic language family (Hebrew). 
We find that, at least in Wikipedia data, a noun's grammatical gender is indeed correlated with the choice of its descriptors.
However, when controlling for a confounder, noun meaning, we present empirical evidence that noun gender has no significant effect on adjective usage.\looseness=-1

\section{A Primer on Grammatical Gender}\label{sec:gender}

In many languages with grammatical gender, adjectives, demonstratives, determiners, and other categories \defn{agree} with the noun in gender, i.e., they will systematically change in form to indicate the grammatical gender of the noun they modify.
Observe the following sentence, \word{A small dog sleeps under the tree.}, translated into two languages that exhibit grammatical gender (German and Polish):\looseness=-1

\begin{enumerate}[label=\alph*., leftmargin=3.4\parindent]
    \item  \normalsize{\word{\textbf{Ein} klein\textbf{er} Hund schl\"aft unter \textbf{dem} Baum.}}  (\textsc{de}) \\
  \small{a.\texttt{M} small.\texttt{M} dog.\texttt{M} sleeps under the.\texttt{M} tree.\texttt{M}}
  
  \item   \normalsize{\word{Mał\textbf{y} pies \'spi pod drzew\textbf{em}.}} (\textsc{pl}) \\
\small{a.\texttt{M} small.\texttt{M} dog.\texttt{M} sleeps under the.\texttt{N} tree.\texttt{N}} 
\end{enumerate}

Because the German (\textsc{de}) and Polish (\textsc{pl}) words for a dog, \word{Hund} and \word{pies}, are both assigned masculine gender, the adjectives in the respective languages, \word{klein} and \word{mały}, are morphologically gender-marked as masculine. 
Additionally, in German, the article, \word{dem}, is also gender-marked as masculine. 
%These are examples of gender concord, the fact that a noun's gender is transparently related to the form of nearby categories, and, according to some (c.f. \citealt{norris-2014-theory, bayirli-2017-universality}, i.a.), noun gender is additionally the driver of concord. 
The fact that gender is reflected by agreement patterns on other elements is generally taken to be a definitional property \citep{hockett-1958-course, corbett1991gender, kramer-2020-grammatical} separating gender from other kinds of noun classification systems, such as numeral classifiers or declension classes.

It is an undeniable fact in many languages that morphological agreement reflects the gender of a noun in the \emph{form} of other elements. 
However, one could imagine a similar process, such as analogical reasoning \citep{lucy-2016-recent}, by which gender could influence both a noun's \emph{meaning} and its form.
If a noun's meaning were to influence its gender, then the noun meaning could also indirectly influence adjective usage, by way of the relationship between grammatical gender and adjective usage. 
There is ample statistical evidence that grammatical gender assignment is not fully arbitrary \citep{williams-etal-2019-quantifying,williams-etal-2021-relationships, nelson2005french, sahai-sharma-2021-predicting}.
Such evidence is \textit{prima facie} consistent with the idea that such influence is conceivably possible.

However, it is important to note that claims that noun gender influences meaning are by their very nature causal claims. 
The most famous example of such a causal claim is the neo-Whorfian view of gender \citep{boroditsky2003sex, boroditsky2003linguistic,Semenuks2017EffectsOG}, which states that a noun's grammatical gender \emph{causally} affects meaning (e.g., adjective choice). 
\citet{boroditsky2003sex} discuss a laboratory experiment they have conducted showing that speakers of German chose stereotypically feminine adjectives to describe, for example, bridges, while speakers of Spanish chose stereotypically masculine adjectives. \citeauthor{boroditsky2003sex} concluded that participant adjective choice reflected the fact that in German, the word for a bridge, \word{Br\"ucke}, is grammatically feminine, while in Spanish, the word for a bridge, \word{puente},
is grammatically masculine.
Their findings can be summed up in the following quote from \citet{boroditsky2003sex}, ``people's ideas about the genders of objects are \emph{strongly influenced} by the grammatical genders assigned to these objects in their native language'' (emphasis ours). Despite this clear causal formulation of the hypothesis, there has yet to be a modeling approach developed to test it.
%Laboratory studies have been used to gather evidence for the neo-Whorfian hypothesis.  For example, \citet{Semenuks2017EffectsOG} perform a small experiment involving human participants to explore whether noun gender affects a particular proxy for meaning, adjective choice. This work found that, in languages where \word{bridge} is feminine (like German; \word{Br{\"u}cke}), participants modified it with adjectives that are stereotypically used to refer to women, such as \word{beautiful}, and in languages where \word{bridge}  is masculine (like Spanish; \word{puente}), they used adjectives stereotypically used to refer to men, like \word{sturdy}.
Moreover, subsequent studies have failed to replicate this result, raising into question the strength of this relationship between gender and adjective usage \citep{mickan2014key} and inviting a study with more appropriate methodology.%\looseness=-1

Our paper builds on \citeposs{williams-etal-2021-relationships} \emph{correlational} study of noun meaning and its distributional properties and advances it to a \emph{causal} one. 
While \citet{williams-etal-2021-relationships} report a non-trivial, statistically significant mutual information between the grammatical gender of a noun and its modifiers, e.g., adjectives that modify the noun, they do not control for other factors which might influence adjective usage, most notably the lexical semantics of the noun. Mutual information is symmetric and thus cannot speak to causation on its own.
We are thus motivated by a potential common cause whereby the lexical semantics jointly influence a noun's grammatical gender \emph{and} its distribution over modifiers and propose a causal model.\looseness=-1

\section{A Causal Graphical Model}\label{sec:model}
The technical contribution of this work is a novel causal graphical model that jointly represents the relationship between the grammatical gender of a noun, its meaning, and descriptors.  
This model is depicted in \Cref{fig:graph}.
If properly estimated, the model should enable us to measure the \emph{causal} effect of grammatical gender on adjective choice in language.
We first develop the necessary notation.\looseness=-1

\paragraph{Notation}
We follow several font and coloring conventions to make our notation easier to digest.
All base sets will be uppercase and in calligraphic font, e.g., $\mathcal{X}$.
Elements of $\mathcal{X}$ will be lowercase and italicized, e.g., $x \in \mathcal{X}$.
Subsets (including submultisets) will be uppercase and unitalicized, e.g., $\mathrm{X} \subset \mathcal{X}$. 
Random variables that draw their values from $\mathcal{X}$ will be uppercase and italicized, e.g., $p(X = x)$.
We will use three colors.
Those objects that relate to nouns will be in {\color{MyBlue} blue}, those objects that relate to adjectives will be in {\color{MyPurple} purple}, and those objects that relate to gender will be in {\color{OliveGreen} green}.\looseness=-1

\subsection{The Model}
\label{sec:the-model}
We assume there exists a set of noun meanings $\nouns$.
In this paper, we assume that such meanings are representable by column vectors in 
$\R^D$.
We denote the elements of $\nouns$ as $\noun \in \R^D$.
Additionally, we assume there exists an alphabet of adjectives $\adjs$.
We denote an element of $\adjs$ as $\adj$.
Finally, we assume there exists a language-dependent set of genders $\genders$.
In Spanish, for instance, we would have $\genders = \{\genfem, \genmsc\}$ whereas in German $\genders = \{\genfem, \genmsc, \genneu\}$.
We denote elements of $\genders$ as $\gender$.\looseness=-1

We now develop a generative model of the subset of lexical semantics relating to adjective choice.
We wish to generate a set of $|\nouns|$ nouns, each of which is modified by a multiset of adjectives.
We can view this model as a partial generative model of a corpus where we focus on generating noun types and adjective tokens.
Generation from the model proceeds as follows:
\begin{align*}
 &\noun \sim \pnoun(\cdot) \\  
&\quad\quad \mathcomment{(\text{sample a noun meaning } \noun)} \\
&\gender_{\noun} \sim \pgender(\cdot \mid \noun) \\ 
&\quad\quad \mathcomment{(\text{sample the gender } \gender_{\noun} \text{ assigned to } \noun)} \\
&\adj_{\noun} \sim \padj(\cdot \mid \noun, \gender_{\noun}) \\
&\quad\quad \mathcomment{(\text{sample adjectives } \adj_{\noun} \text{ that modify } \noun)} 
\end{align*}
In this formulation, $\rvN$ is a $\nouns$-valued random variable, $\rvG$ is a $\genders$-valued random variable, and $\rvA$ is a $\adjs$-valued random variable.\looseness=-1

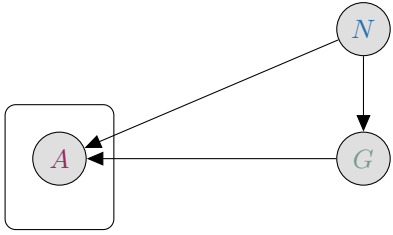
\begin{figure}
  \centering
  \begin{tikzpicture}

  % Define nodes

  \node[obs, yshift=-1cm] (a) {$\rvA$};
  \node[obs, above = of a, xshift=4cm]  (n) {$\rvN$};
  \node[obs, below=of n]  (g) {$\rvG$};
  % \node[latent, above = of a, xshift=-2.5cm] (A) {$\adjmeans$};
 % \node[latent, right=of n] (mu) {$\vmu$};
 % \node[latent, below=of mu] (sigma2) {$\sigma^2$};

  % Connect the nodes
  \edge {n,g} {a};%
  \edge {n} {g};

  % Plates
  %\plate {yx} {(x)(y)} {$N_2$} ;
  % \plate {} {(w)(y)x(yx.north west)(yx.south west)} {$N_1$} ;
  %\plate [inner sep=0.55cm, xshift=-0.3cm, yshift=-0cm, minimum width=\columnwidth] {} {(a)(g)(n)} {$1 \leq i \leq \nnoun$} ;
    \plate [inner sep=0.35cm]{} {(a)} {}
  \end{tikzpicture}
  \caption{Causal graphical model relating noun semantics, gender, and adjective choice.
  The neo-Whorfian hypothesis posits that a noun's gender \emph{causally} influences adjective choice.
  Correctly evaluating this hypothesis must also account for the relationship between the noun's meaning and adjective choice.\looseness=-1
  }
  \label{fig:graph}
  \vspace{-10pt}
\end{figure}

Written as a probability distribution,
we have
\begin{align}
\label{eq:gen_model}
p(&\{\setA_{\noun}\}, \{\gender_{\noun}\}, \setN) \\
&= \prod_{\noun \in \setN} \prod_{\adj \in \setA_{\noun}}\,\padj(\adj \mid \noun, \gender_{\noun})\, \pgender(\gender_{\noun} \mid \noun)\,\pnoun(\noun) \nonumber
\end{align}
where $\setN \subset \nouns$ is a subset of the set of noun meanings and each $\gender_\noun \in \genders$ is the gender of $\noun$, and each $\setA_{\noun} \subset \adjs$ is a multisubset of $\adjs$ that contains the observed adjectives that modify $\noun$.
This model is represented graphically in \Cref{fig:graph}, where the arrow from $\rvN$ to $\rvG$ represents the dependence of $\rvG$ on $\rvN$ as shown in the conditional probability distribution $\pgender(\gender_{\noun} \mid \noun)$, and the arrows from $\rvN$ and $\rvG$ to $\rvA$ represent the dependence of $\rvA$ on $\rvN$ and $\rvG$, as shown in the conditional probability distribution $\padj(\adj \mid \noun, \gender_{\noun})$.\looseness=-1

Importantly, our model \emph{generates} the lexical semantics of noun types.
This means that a sample from it generates a new noun, whose semantics we may never have seen before.
If we are able to estimate such a model well, we can use the basics of causal inference 
to estimate the causal effect gender has on adjective usage.
Specifically, as is clear from \Cref{fig:graph}, the only confounder between gender and adjective selection in our proposed model is the semantics of the noun.\footnote{Sentential context can also influence adjective usage, e.g., the probability distribution over adjectives describing the noun \word{bagel} might differ between the sentences \word{After the flood, the rat discovered a \_\_\_ bagel dissolving in the sewer.}, and \word{She was craving a \_\_\_ bagel.} 
Our model does not aim to account for such contextual effects.}

\subsection{Intervention}
Thus, to the extent that the modeler believes our model $p$ is a reasonable generative model of lexical semantics, we apply Pearl's backdoor criterion to get a causal effect \citep{pearl1993bayesian}.
One does so by applying the $\mathrm{do}$-calculus, which results in the following gender-specific distribution over adjectives\looseness=-1
\begin{align}
\label{eq:backdoor}
p(\adj \mid &\,\mathrm{do}(\rvG = \gender)) \\
&= \sum_{\noun \in \nouns} \padj\left(\adj \mid \rvG = \gender, \noun\right) \pnoun(\noun)  \nonumber
\end{align}
where for simplicity, $\nouns$ is assumed to be at most countable, despite being a subset of $\R^D$.
We are now interested in using $p(\adj \mid \mathrm{do}(\rvG = \gender))$ to measure the extent to which the grammatical gender of a noun influences which adjectives are used to modify that noun. 
In particular, we aim to measure how different the adjective choice would be if the noun had a different grammatical gender.
Because $p(\adj \mid \mathrm{do}(\rvG = \gender))$ is a distribution over $\adjs$,
we measure the causal effect by the weighted Jensen--Shannon divergence \citep{jensenshannon}, which we define as
\begin{equation}
\begin{split}
\JSD_{\pi}&(p_1 \mid\mid p_2) \\
& \defeq \pi_1 \KL(p_1 \mid\mid m) + \pi_2 \KL(p_2 \mid\mid m) 
\end{split}
\end{equation}
where $\pi_1, \pi_2 \geq 0$, $\pi_1 + \pi_2 = 1$ and
$m = \pi_1 p_1 + \pi_2 p_2$ is a convex combination of $p_1$ and $p_2$ weighted according to $\pi$.\footnote{The Jensen--Shannon divergence can also be generalized to operate on $N$ distributions as $\JSD_{\pi}(p_1, \ldots, p_N) = \sum_{n=1}^N \pi_n\KL(p_n \mid\mid m)$, where $\sum_{n=1}^N \pi_n = 1$, $\pi_n \geq 0,\,\, \forall n \in [N]$, and $m = \sum_{n=1}^N \pi_n p_n$.} 
Further, we note that the weighted Jensen--Shannon divergence is related to a specific mutual information between two random variables. 
We make this relationship formal in the following proposition.

\begin{restatable}{prop}{jsmi}
\label{prop:jsmi}
Let $\rvA$ and $\rvG$ be $\adjs$-valued and $\genders$-valued random variables, respectively.
Further assume they are jointly distributed according to $p(\adj \mid \mathrm{do}(\rvG=\gender)) \pgender(\gender)$.
Then, 
\begin{equation}
    \JSD_{\pgender}\Big(\Big\{p(\cdot \mid \mathrm{do}(\rvG=\gender))\Big\}\Big) = \MIdo(\rvA;\rvG)
\end{equation}
where $\MIdo(\rvA; \rvG)$ is the mutual information computed under the joint distribution $p(\adj \mid \mathrm{do}(\rvG=\gender))\,\pgender(\gender)$.\looseness=-1
\end{restatable}

\begin{sproof}
See \Cref{sec:proof} for a proof.
\end{sproof}

Relating the weighted Jensen--Shannon divergence to a specific mutual information provides a clear interpretation. 
This measure explains in bits how much the entropy of the language's distribution over adjectives is reduced when the grammatical gender of the noun being modified is known at the time of the adjective choice. 
For instance, if the language's distribution over adjectives has an entropy $\mathrm{H}(\rvA)$ of 10 bits and the mutual information $\MI(\rvA; \rvG) \defeq \mathrm{H}(\rvA) - \mathrm{H}(\rvA \mid \rvG)$ is 1 bit, then knowing the gender allows us to reduce the uncertainty over which adjectives modify the nouns to $\mathrm{H}(\rvA \mid \rvG) = 9$ bits.
However, the reduced uncertainty measured by $\MI(\rvA; \rvG)$ is purely associational; we cannot conclude that the gender of the noun actually causes the change in adjective distribution. Such a change could also be attributed to a confounding factor (like noun meaning).
On the other hand, $\MIdo(\rvA; \rvG) \defeq \Hdo(\rvA) - \Hdo(\rvA \mid \rvG)$ represents the amount of uncertainty in the adjective distribution \emph{causally} reduced by the gender random variable.
Intuitively, we can reason about $\Hdo(\rvA)$ and $\Hdo(\rvA \mid \rvG)$ as the uncertainty of the adjective distribution in a world where we can counterfactually imagine that all nouns have the same gender $\gender$, and thus by setting all else equal, isolate the effect of knowing gender alone on the uncertainty of the adjective distribution.
For a formal definition of $\Hdo(\rvA)$ and $\Hdo(\rvA \mid \rvG)$, see  \Cref{sec:proof}.\looseness=-1

\subsection{Intuition}\label{sec:intuition}
We now explain the intuition behind the mutual information $\MIdo(\rvA; \rvG)$.
%using a toy example. 
%Consider a language with a set of adjectives $\adjs = \{ \wordadj{awful}, \wordadj{sweet}, \wordadj{feminist}\}$, a set of nominal meanings $\nouns = \{ \wordnoun{bridge}, \wordnoun{cake}, \wordnoun{ship}\}$, vectors in $\R^D$, 
%and grammatical gender $\genders = \{\genfem, \genmsc, \genmsc\}$.\looseness=-1

\paragraph{Case 1: No edge from $\rvN$ to $\rvG$.}
First, consider the case when there is no edge from $\rvN$ to $\rvG$, indicating that there is no causal relationship between a noun's meaning and its grammatical gender. 
Under this condition, we have $\pgender(\gender \mid \noun) = \pgender(\gender)\ \forall \gender \in \genders, \noun\in \nouns$.
Consequently, the interaction between the grammatical gender of a noun and the adjectives used to describe this noun is not mitigated by the meaning of this noun, and thus, $\MIdo(\rvA; \rvG) = \MI(\rvA; \rvG)$.\looseness=-1

\paragraph{Case 2: No edge from $\rvG$ to $\rvA$.}
Second, consider the case when there is no edge from $\rvG$ to $\rvA$, indicating that there is no causal relationship between a noun's gender and its adjective distribution.
In this case, $\MIdo(\rvA; \rvG) = 0$. Further, we can show that $\MIdo(\rvA; \rvG) = 0$ if and only if the edge from $\rvG$ to $\rvA$ does not exist.

\paragraph{Case 3: All edges.} 
Finally, when both edges from $\rvG$ to $\rvN$ and $\rvG$ to $\rvA$ exist, $\MIdo(\rvA; \rvG)$ can vary. 
In particular, $\MIdo(\rvA; \rvG)$ is non-negative (and indeed non-zero by case 2).
However, we know of no relationship between $\MIdo(\rvA; \rvG)$ and $\MI(\rvA; \rvG)$. 
In this case, the strength of the relationship between a noun's grammatical gender and adjective choice is regulated by the meaning.\looseness=-1

\subsection{Parameterization}\label{sec:parameterization}

We now discuss the parameterization of the conditional distributions given in \Cref{sec:the-model}: adjectives ($\padj$), gender ($\pgender$), and vector representations of nouns ($\pnoun$).
We model $\padj$ using a logistic classifier where the probability of each adjective $\adj$ is predicted given the column vector $\left[\embed(\adj); \noun; \embed(\gender) \right]$,
which is a stacking of a column vector representation $\embed(\adj)$ of an adjective $\adj$, the meaning representation of the noun $\noun$, and a column vector representation of gender $\gender$, respectively. 
The classifier's functional form is given as
\looseness=-1
\begin{align}
\label{eq:p_adj_given_everything}
  \padj(&\adj \mid \gender, \noun) \\
  &= \frac{\exp\left(  \vecw^{\top}\tanh \boldsymbol{W} \left[\embed(\adj); \noun; \embed(\gender)  \right] \right) }{\sum_{\adjtwo \in \adjs} \exp\left( \vecw^{\top}\tanh \boldsymbol{W} \left[\embed(\adjtwo); \noun; \embed(\gender) \right] \right)} \nonumber
\end{align}
where the parameters $\boldsymbol{W}$ and $\vecw$ denote the weight matrix and weight column vector, respectively.
We note that \Cref{eq:p_adj_given_everything} gives the probability of a single $\adj \in \setA_{\noun}$ that co-occurs with $\noun$.
The probability of the set $\setA_{\noun}$ is the product of generating each adjective independently. 
While $\embed(\adj)$ and $\embed(\gender)$ could be trainable parameters, for simplicity, we fix $\embed(\adj)$ to be standard word2vec representations and $\embed(\gender)$ to be a one-hot encoding with dimension $|\genders|$.
Representations for $\noun$ are pre-trained according to methods described in \Cref{sec:word_embs}.

We opt to model $\pgender(\gender \mid \noun)$ and $\pnoun(\noun)$ as the empirical distribution of nouns in the corpus.\looseness=-1

\begin{table*}[ht]
    \centering
    \fontsize{10}{10}\selectfont

    \begin{tabular}{lrrrrr}
        \toprule
          & \textsc{de} & \textsc{es} & \textsc{he} & \textsc{pl} & \textsc{pt} \\ \midrule 
          word2vec & & & & & \\ \midrule
         \spacebefore \# noun types & 932 & 953 & 814 & 891 & 929 \\
         \spacebefore \# adj types & 109,549 & 61,839 & 29,855 & 42,271 & 30,004 \\
         \spacebefore \# noun-adj types & 486,647 & 581,589 & 208,202 & 223,774 & 176,995 \\
         \spacebefore \# noun-adj tokens & 5,966,400 & 7,523,601 & 2,413,546 & 4,040,464 & 1,543,563 \\ \midrule
         WordNet & & & & & \\ \midrule
         \spacebefore \# noun types & 437 & 773 & 391 & 450 & 630 \\
         \spacebefore \# adj types & 78,585 & 58,536 & 26,278 & 38,427 & 26,112 \\
         \spacebefore \# noun-adj types & 272,511 & 513,905 & 145,542 & 178,049 & 134,923 \\
         \spacebefore \# noun-adj tokens & 3,606,909 & 6,912,761 & 1,978,561 & 3,493,547 & 1,243,506 \\
        \bottomrule
    \end{tabular}
    
    \caption{Data statistics in our Wikipedia corpora with retrieved word2vec and WordNet representations.}
    \label{tab:data}
    \vspace{-10pt}
\end{table*}

\section{Experimental Setup}
\label{sec:experimental}

In this section, we describe the data used in our experiments, and how we estimate non-contextual word representations as a proxy for a noun's lexical semantics.\looseness=-1

\subsection{Data} 
\label{sec:data}
We gather data in five languages that exhibit grammatical gender agreement: German, Hebrew, Polish, Portuguese, and Spanish. 
This is certainly not a representative sample of the subset of the world's languages that exhibit grammatical gender, but we are limited by the need for a large corpus to estimate a proxy for lexical meaning.
Hebrew, Portuguese, and Spanish distinguish between two grammatical genders (masculine and feminine), while German and Polish distinguish between three genders (masculine, feminine, and neuter).\footnote{Polish also includes an animacy distinction for masculine nouns.}\looseness=-1

We use the Wikipedia dump dated August 2022 to create a corpus for each of the five languages,\footnote{\url{https://dumps.wikimedia.org/}} and preprocess the corpora with the Stanza library \citep{qi-etal-2020-stanza}.\footnote{\url{https://stanfordnlp.github.io/stanza/}} 
Specifically, we tokenize the raw text, dependency parse the tokenized text, lemmatize the data, extract lemmatized noun--adjective pairs based on an \texttt{amod} dependency label, and finally filter these pairs such that only those for inanimate nouns remain.
To determine which nouns are inanimate, we use the NorthEuraLex dataset, a curated list of cross-linguistically common inanimate nouns \citep{dellert2020lang}. 
\Cref{tab:data} shows the counts for the remaining tokens for all analyzed languages for which we retrieved word representations. 
Next, we describe the procedure for computing the non-contextual word representations. 

\subsection{Non-contextual Word Representations}
\label{sec:word_embs}

%In \Cref{sec:model}, we described a causal graphical model of the interactions between a noun's meaning, grammatical gender, and adjectives.
The model described in \Cref{sec:model} relies on a representation of nominal lexical semantics---specifically, a representation independent (in the probabilistic sense) of the distributional properties of the noun.\footnote{We describe two ways in which we construct such representations.
Similar to this approach, \citet{kann2019grammatical} trains a classifier to predict gender from word representations trained on a lemmatized corpus.}\looseness=-1 %However, they fail to find an effect unlike \citet{williams-etal-2021-relationships}.

\paragraph{Word2vec.}

We train word2vec \citep{mikolov2013word} on modified Wikipedia corpora. 
As found in \citet{sabbaghi2022measuring}, word representations learn the association between
a noun and its grammatical gender in grammatically gendered languages.
Thus, we first lemmatize the corpus with Stanza as discussed above.
This step should remove any spurious correlations between a noun's morphology and its meaning.
Second, we remove all adjectives from the corpora.
Because our goal is to predict the distribution over adjectives \emph{from} a noun's lexical semantic representation, that distribution should not, itself, be encoded in the semantic representation.
We construct representations of length 200 through the continuous skip-gram model with negative sampling with 10 samples using the implementation from \texttt{gensim}.\footnote{\url{https://radimrehurek.com/gensim/models/word2vec.html}} We train these non-contextual word representations on the Wikipedia data described above. We ignore all
words with a frequency below five and use a symmetric context window size of five.\looseness=-1

\paragraph{WordNet-based Representations.}
In addition to those representations derived from word2vec, we also derive lexical representations using WordNet \citep{miller-1994-wordnet}.
%Specifically, we follow \citeposs{saedi-etal-2018-wordnet} procedure.
Because WordNet is a lexical database that groups words into sets of synonyms (synsets) and links synsets together by their conceptual, semantic, and lexical relations, representations of meaning based on WordNet are unaffected by biases that might be encoded in a training corpus of natural language.
Following \citet{saedi-etal-2018-wordnet}, we create word representations by constructing an adjacency matrix of WordNet's semantic relations (e.g., hypernymy, meronymy) between words and compressing this matrix to have a dimensionality of 200 for each of the languages in this study: German, Hebrew, Spanish, Polish, and Portuguese \citep{siegel-bond-2021-odenet, ordan2007hebrewwordnet, gonzalez2013spanishwordnet, Piasecki2009polishwordnet, Paiva2012portuguesewordnet}.
We access and process these WordNets using the Open Multilingual WordNet \citep{bond2012survey}.
We report statistics on these WordNets in \Cref{tab:wordnetstats}.

\begin{table}
\centering
\fontsize{10}{10}\selectfont
\begin{tabular}{@{}lrrr@{}}
\toprule
WordNet                                   & Words           & Senses & Synsets \\ \midrule
ODENet 1.4 (de)              & 120,107 & 144,488 & 36,268   \\
OpenWN-PT (pt)                    & 54,932           & 74,012  & 43,895   \\
plWordNet (pl)                        & 45,456           & 52,736  & 33,826   \\
MCR (es) & 37,203           & 57,764  & 38,512   \\
% Arabic WordNet (ar)                            & 18,003           & 37,342  & 9,962    \\
Hebrew WordNet (he)                           & 5,379            & 6,872   & 5,448    \\ \bottomrule
\end{tabular}

\caption{
    Summary statistics on the WordNets used for training representations in each language.
}
\label{tab:wordnetstats}
  \vspace{-10pt}
\end{table}

\paragraph{Evaluating the Representations.}
We now discuss how we validate our lexical representations.
Because we construct the word2vec representations using modified corpora, it is reasonable to fear that those modifications would hinder the representations' ability to encode an adequate approximation to nominal lexical semantics.
Thus, for each language, we evaluate the quality of the learned representations by calculating the Spearman correlation coefficient of the cosine similarity between representations and the human-annotated similarity scores of word pairs in the SimLex family of datasets \citep{hill-etal-2015-simlex, leviant2015msimlex999, vulic2020multisimlex}.
A higher correlation indicates a better representation of semantic similarity.
We report the Spearman correlation of the representations for each language in \Cref{tab:wordrepresentationssimilarity}.
Especially for representations generated using WordNet for languages with sparsely-populated WordNets (see \Cref{tab:wordnetstats}), the representational power is relatively low (as measured by the Spearman correlation), which may influence conclusions of downstream results for these languages. 
We note that if the representations are very bad---such that gender is completely unpredictable from the noun meaning representation and $\pgender(\gender_{\noun} \mid \noun) = \pgender(\gender_{\noun})$---then $\MI(\rvA; \rvG) = \MIdo(\rvA; \rvG)$ because the edge in the graphical model from $\rvN$ to $\rvG$ is effectively removed.\looseness=-1

\section{Methodology}
We now outline the methodology of our study, starting with parameter estimation of the model in \cref{sec:estimation} and plug-in estimation of $\MI(\rvA; \rvG)$ in \cref{sec:plug-in}. 
We conduct two experiments: first, for a point of comparison, we replicate \citeposs{williams-etal-2021-relationships} study to estimate $\MI(\rvA; \rvG)$ for each of the five languages (\cref{sec:exp1});
and second, we produce a causal analog of \citet{williams-etal-2021-relationships} (\cref{sec:exp2}).
Using the notation of \cref{sec:model}, in the second experiment, we estimate $\MIdo(\rvA; \rvG)$ for each of the five languages. Finally, in \cref{sec:permutation-testing}, we discuss our permutation testing methodology.\looseness=-1

\subsection{Parameter Estimation}\label{sec:estimation}
To estimate the parameters of the graphical model given in \cref{fig:graph}, we perform regularized maximum-likelihood estimation.
Specifically, we maximize the likelihood the model assigns to a training set $\Dtrn= \{(\setA_n, \gender_n, \noun_n)\}_{n=1}^N$ where each distinct $\noun_n$ occurs at most once.
The log-likelihood is
\begin{align}
\mathcal{L}(\vtheta) = \sum_{n=1}^N \sum_{\adj \in \setA_n} \log \padj(\adj \mid \gender_n, \noun_n) 
\end{align}
where $\vtheta = \{\vecw, \boldsymbol{W} \}$.
We define $\padj$ using a multilayer perceptron (MLP) with the rectified linear unit~\citep[ReLU;][]{nairRectifiedLinearUnits2010}
and a final softmax layer.
We estimate $\pnoun$ and $\pgender$ from the empirical distributions derived from the corpus. We further apply $L_{1}$-regularization to impose sparsity and $L_{2}$-regularization to prevent a representation's dimension from dominating the model's predictions. 
The regularization coefficients are each set to 0.001.
We train our models for each of the five languages for a maximum of $100$ epochs using the Adam optimizer~\citep{kingmaAdamMethodStochastic2015} to predict the adjective given its representation, a noun's gender, and representation.%\looseness=-1 

\begin{table}
    \centering
    \fontsize{10}{10}\selectfont
    \begin{tabular}{@{}lrrrr@{}}
    \toprule
     & \multicolumn{2}{c}{WordNet} & \multicolumn{2}{c}{word2vec} \\
    \cmidrule(lr){2-3} \cmidrule(lr){4-5}
     Lang. & $\rho$ & \% of eval set & $\rho$ & \% of eval set \\
    \midrule 
    \textsc{de} & 0.360 & 86.9\% & 0.380 & 92.2\% \\
    \textsc{es} & 0.234 & 71.8\% & 0.419 & 89.3\% \\
    \textsc{he} & 0.104 & 11.6\% & 0.460 & 59.6\% \\
    \textsc{pl} & 0.092 & 49.9\% & 0.418 & 76.5\% \\
   \textsc{pt} & 0.283 & 94.7\% & 0.308 & 94.5\% \\
    \bottomrule
\end{tabular}
    \caption{
    Spearman's $\rho$ correlation coefficient between judgments in similarity datasets and representation cosine similarity for each language for both WordNet and word2vec representations.}
\label{tab:wordrepresentationssimilarity}
\vspace{-15pt}
\end{table}

\subsection{Plug-in Estimation of $\MI(\rvA; \rvG)$}\label{sec:plug-in}
The first estimator of $\MI(\rvA; \rvG)$ is the plug-in estimator considered by \citet{williams-etal-2021-relationships}. 
In this case, we compute the maximum-likelihood estimate (MLE) of the marginal $p(\adj, \gender)$ and plug it into the formula for mutual information:\footnote{While we opt for the MLE approach to maintain consistency with \citet{williams-etal-2021-relationships}, we note that the alternative entropy estimators instead might have yielded lower mutual information estimates, as suggested by \citet{arora-etal-2022-estimating}.}  
\begin{equation}
\begin{split}
\label{eq:mi}
    \MI(\rvA; \rvG) = \sum_{\adj \in \adjs} \sum_{\gender \in \genders} p(\adj, \gender) \log \frac{p(\adj,\gender)}{p(\gender)p(\adj)}
\end{split}
\end{equation}
Following \citet{williams-etal-2021-relationships}, we use empirical probabilities as the plug-in estimates.

\subsection{Model-based Estimation of $\MI(\rvA; \rvG)$}
\label{sec:exp1}
In the first experiment, we replicate \citeauthor{williams-etal-2021-relationships}'s findings on different data and with a different method.
Let $p(\adj, \gender, \noun) = \padj(\adj \mid \gender, \noun) \pgender(\gender \mid \noun) \pnoun(\noun)$ be an estimated
model that factorizes according to the graph given in \cref{fig:graph}, and let $\nounseval$ be a set of gender--noun pairs where the nouns are \emph{distinct} from those in the test set, $\Dtst$. Let $\gendertwo$ and $\nountwo$ be gender--noun pairs from this test set. 
Using $\nounseval$, consider the following approximate marginal: 
\begin{equation}
\widetilde{p}(\adj, \gender) = \frac{1}{|\nounseval|}\sum_{(\gendertwo, \nountwo) \in \nounseval} \!\!\!\!\padj(\adj \mid \gendertwo, \nountwo) \mathbbm{1}\{\gender = \gendertwo\} 
\end{equation}%
We then plug $\widetilde{p}(\adj, \gender)$ into the formula for correlational $\MI(\rvA; \rvG)$ defined in \cref{eq:mi}. 
%We perform a permutation test to test whether the estimate is significantly different than zero.
%The method is described in \cref{sec:permutation-testing}.

\subsection{Model-based Estimation of $\MIdo(\rvA; \rvG)$} 
\label{sec:exp2}
Here, in contrast to \cref{sec:exp1}, we are interested in \emph{causal} mutual information, which we take to be
the mutual information as defined under $p(\adj \mid \mathrm{do}\left(\rvG = \gender\right)) p(\gender)$.
We approximate the marginal $p(\gender)$ using a maximum-likelihood estimate on $\Dtrn$. We use $\nounseval_{\gender}$, a set of gender--noun pairs \emph{distinct} from those in $\Dtst$ with a fixed gender $\gender$ to compute the following estimate of the intervention distribution
\begin{equation}
\begin{split}
\widetilde{p}(\adj \mid \,&\mathrm{do}(\rvG= \gender)) \\
&=\frac{1}{|\nounseval_{\gender}|} \sum_{(\gender, \nountwo) \in \nounseval_{\gender}} \padj(\adj \mid \gender,  \nountwo) 
\end{split}
\end{equation}
using the parameters of the model $\padj(\adj \mid \rvG=\gender, \noun)$ estimated as described in \cref{sec:estimation}.
We use a permutation test to determine whether the estimate is significantly different than zero, as described in \cref{sec:permutation-testing}.\looseness=-1

\subsection{Permutation Testing}\label{sec:permutation-testing}
To do this, we train a model from scratch using 5-fold cross-validation on a subset of 100 adjectives to estimate $\padj(\adj \mid \noun, \gender)$ with a random permutation of the gender labels and use that model to compute the pair-wise mutual information estimates between adjective distributions on the test set as described earlier for $k=2,000$ times (a total of 10,000 runs). 
We design and run a permutation test to determine whether the mutual information between the adjective distributions conditioned on different genders is equal to the mutual information between the adjective distributions from a model trained on perturbed gender labels.
We determine the significance of our result by evaluating the proportion of times that $\MIdo(\rvA; \rvG)$, as computed using the non-permuted training set, is lower than one computed using randomly permuted genders during training.
We choose the standard significance level of $\alpha=.05$; that is, when we observe a $p$-value lower than $.05$, we reject the null hypothesis, which posits no difference in mutual information between models trained on original and perturbed gender labels.

\begin{figure}
    \centering
    \includegraphics[width=\columnwidth]{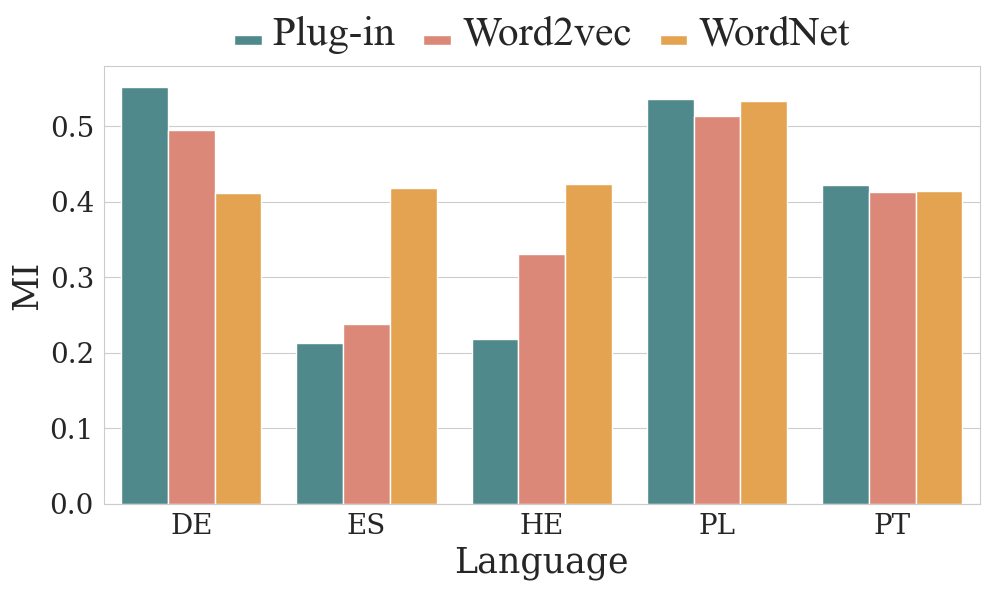}
    \caption{Results for the plug-in estimation of $\MI(\rvA; \rvG)$ and model-based estimations for $\MI(\rvA; \rvG)$. % using the word2vec and WordNet representations.
    }
    \label{fig:model-valid}
\end{figure}

\begin{table*}

    \centering
    \fontsize{10}{10}\selectfont
%\begin{adjustbox}{max width=\textwidth}

    \begin{tabular}{lrrrrrr}
        \toprule
        & \multicolumn{3}{c}{word2vec} & \multicolumn{3}{c}{WordNet} \\ \cmidrule(lr){2-4} \cmidrule(lr){5-7}
        & Model-based & Model-based& Mean diff. & Model-based & Model-based & Mean diff. \\
         Language & $\MI(\rvA; \rvG)$ & $\MIdo(\rvA; \rvG)$ & Perturbed & $\MI(\rvA; \rvG)$ & $\MIdo(\rvA; \rvG)$ & Perturbed \\ 
        \midrule 
        \textsc{de} & 0.526 & $\expnumber{1.24}{-4}$ & $\expnumber{3.12}{-4}^{*}$& 0.412 &$\expnumber{2.17}{-5}$ & $\expnumber{1.03}{-2}^{*}$  \\
        \textsc{es} & 0.238 & $\expnumber{4.60}{-5}$ & $\expnumber{4.85}{-4}^{*}$ & 0.418 & $\expnumber{1.24}{-5}$& $\expnumber{1.77}{-3}^{*}$ \\
        \textsc{he} & 0.331 & $\expnumber{8.03}{-4}$ & $\expnumber{4.70}{-3}^{*}$& 0.423 & $\expnumber{1.43}{-5}$& $\expnumber{1.11}{-3}^{*}$  \\
        \textsc{pl} & 0.545 & $\expnumber{1.65}{-4}$ & $\expnumber{6.67}{-4}^{*}$& 0.533 & $\expnumber{8.68}{-7}$& $\expnumber{1.37}{-4}^{*} $ \\
        \textsc{pt} & 0.413 & $\expnumber{1.72}{-4}$ &$\expnumber{6.31}{-3}^{*}$ &  0.414 & $\expnumber{8.80}{-5}$& $\expnumber{1.76}{-3}^{*}$ \\ \bottomrule
    \end{tabular}
    
 %   \end{adjustbox}
    \caption{Results for the plug-in estimation of $\MI(\rvA; \rvG)$, model-based estimation for $\MI(\rvA; \rvG)$, and model-based estimation of $\MIdo(\rvA; \rvG)$, mean difference between the model-based estimation of $\MIdo(\rvA; \rvG)$ and a perturbed model with random gender labels for the causal model trained with word2vec and WordNet representations.
    %Mutual information (MI) between gender and adjective choice and Jensen--Shannon divergence (JSD) for the causal graphical model and mean difference (MD) in JSD to the perturbed model together with 
    Significant differences ($p$-value $<.05$) according to the permutation test are marked with an asterisk.
    }
    \label{tab:results-adj}
   \vspace{-10pt}
\end{table*}

\section{Results}
\label{sec:results}
%In this section, we present the results of our empirical investigation. 
%First, we examine whether noun meaning influences the gender of that noun (\S\ref{sec:res-gender}). 
First, we validate our model by comparing the model-based estimation of $\MI(\rvA; \rvG)$ to the method presented in \citet{williams-etal-2021-relationships}, the plug-in estimation of $\MI(\rvA; \rvG)$.
Then, we employ our causal graphical model to investigate whether there is evidence for the neo-Whorfian claim that the grammatical gender of a noun influences the adjective chosen to modify this noun, even when we control for the meaning of the noun.

%\subsection{Grammatical Gender Assignment}
%\label{sec:res-gender}

%In \Cref{tab:results-gender}, we present the performance of the grammatical gender classification given the meaning of the noun for each of the analysed languages. Both models trained with word2vec embeddings and WordNet embeddings clearly outperform the baseline classifier in terms of the $\Fscore$-score for all analysed languages.  This result indicates that without controlling for the noun meaning, prior studies might have overestimated the influence of gender on the adjective choice. We note that except for Portuguese, the model trained with word2vec embeddings outperforms the model trained with WordNet embeddings suggesting that word2vec embeddings are potentially better at representing lexical semantics.  

%\subsection{The Neo-Whorfian Hypothesis}
%\label{sec:res-whorfian}
We first validate our model by comparing its results to \citeposs{williams-etal-2021-relationships} plug-in estimate of $\MI(\rvA;\rvG)$. 
If the results of both of these estimates are comparable, we have evidence that our model indeed captures the relation between grammatical gender and adjective choice. 
We present the results in \cref{fig:model-valid}. 
We observe a substantial relationship between grammatical gender and adjective usage based on the plug-in and model-based $\MI(\rvA;\rvG)$ estimates replicating the results of \citet{williams-etal-2021-relationships}.
The estimates of the model-based $\MI(\rvA;\rvG)$ computed using both word2vec and WordNet representations, and the plug-in $\MI(\rvA;\rvG)$ lie between 0.2 and 0.5, with the estimates of the model-based approach being consistently higher (with the exception of German) than the estimates of the plug-in $\MI(\rvA;\rvG)$. 
Thus, the non-zero estimates of the model-based $\MI(\rvA;\rvG)$ indicate that some relationship exists between a noun's grammatical gender and adjective usage.\looseness=-1 

Given the above result, we are interested in whether the strength of this relationship is mitigated when controlling for the meaning of a noun. 
We present the estimates of the model-based $\MIdo(\rvA;\rvG)$ in \Cref{tab:results-adj} and compare them to the model-based estimates of the $\MI(\rvA;\rvG)$.
While we observe evidence for the influence of grammatical gender on adjective choice in a non-causal setup based on $\MI(\rvA;\rvG)$, this relationship shrinks to close to 0 when we control for noun meaning in our causal model trained using both word2vec and WordNet representations. 
%Additionally, the difference between the causal MI of our model and a model trained on perturbed gender labels is roughly 0 and is not statistically significant. 

For completeness, we test for the presence of a difference between the size of the $\MIdo(\rvA;\rvG)$ of our model and a model trained on randomly perturbed gender labels based on a subset of adjectives. We reject the null hypothesis that the distributions are exactly the same for all languages and representations' settings. %Thus, we infer that in the 
%Further, for the two languages with three gender classes (German and Polish), we investigate the difference between the mutual information of the distribution over adjectives for neuter gender and masculine and feminine gender, respectively. While the difference between neuter and masculine is $\expnumber{6.11}{-9}$ for German and 0 for Polish, for the feminine gender it is visibly higher at $\expnumber{1.87}{-4}$ for German and $\expnumber{2.84}{-4}$ for Polish.

\section{Discussion}
% Our main finding
\paragraph{Evidence against the neo-Whorfian hypothesis.}
We find that the interaction between the grammatical gender of inanimate nouns and the adjectives used to describe those nouns all but disappears when controlling for the meaning of those nouns, for all five analyzed gendered languages. 
While the order of magnitude of $\MIdo(\rvA;\rvG)$ measured with our model is significantly different from that of a model trained on random gender labels, it remains minuscule in absolute terms.
This minor difference points towards the absence of a meaningful causal relationship between a noun's gender and its modifiers in the languages studied.
% For completeness, we conduct significance testing and find that we fail to reject the null hypothesis that the distributions are the same.
Thus, we provide an additional piece of evidence against the neo-Whorfian hypothesis in noun--adjective patterns.\looseness=-1
% The question isn't whether there is any difference, but whether that difference is meaningful. The most important consideration here is the size of the effect, which we quantify with mutual information. We find that the order of magnitude of the measured causal MI is miniscule, suggesting there is no meaningful causal relationship between noun meaning (as we operationalize it), and gender assignemnt in the languages studied. For completeness, we also report the p-values, and find that we fail to reject the null hypothesis that the distributions are the same. "We failed to find evidence that the distributions are the same (JSD p-value), evidence is consistent with no difference between the distributions"

% another weakening to the neo-whorfian hypothesis
\paragraph{A possible weakening of the neo-Whorfian hypothesis.} Although the size of the overall effect is small, it is possible that the effect of gender on adjective choice is stronger for some words than others.
Future work could explore whether there is evidence of a noticeable effect of gender on adjective choice for a more restricted set of inanimate nouns, e.g., referring to artifacts or body parts. Such evidence could perhaps support a weakened version of the neo-Whorfian hypothesis.

% Discuss that wordnet has lower MI consistently than word2vec. Speculate that this is caused by word2vec being not completely biased because gender might still be encoded from the surrounding verbs, pronouns, etc., while wordnet doesn't have such bias. 
\paragraph{Comparing results between word2vec and WordNet.}
Our results hold for both of the word representation conditions, word2vec, and WordNet. 
Notably, in comparing the two, we find that using WordNet representations consistently results in a lower $\MIdo(\rvA; \rvG)$ than word2vec for all languages analyzed in this study.
One possible explanation for this difference is that, despite our efforts to make non-contextual word2vec representations, these word2vec representations may still encode some signal regarding gender from the remaining context (such as verb choice or adjacent gendered pronouns in the corpora).
If these word2vec representations contain unwanted context-based gender information in addition to the noun meaning, it could result in overestimating $\MIdo(\rvA; \rvG)$.
Furthermore, since WordNet representations are created independently from any context within a corpus, they should not contain any signal related to grammatical gender, which may therefore be reflected in the consistently lower $\MIdo(\rvA; \rvG)$.

\paragraph{Design choices and limitations.} 
We note several choices in the experimental setup
which may influence this analysis.
First, while we experiment with NorthEuraLex, which furnishes us with a list of inanimate nouns, the dataset excludes rarer nouns for which an effect might be observed.\footnote{We note that the original laboratory experiments taken to be as evidence for the neo-Whorfian hypothesis \citep{boroditsky2003sex, Semenuks2017EffectsOG} also only used high-frequency nouns. Moreover, if an effect were observed mainly with respect to low-frequency nouns, this would further weaken the neo-Whorfian hypothesis.} 
Second, while non-contextual word representations are the current de facto proxy for lexical semantics, they remain a proxy and are fundamentally limited.
Furthermore, in our effort to estimate word2vec representations for noun meaning without encoding gender-based context, we chose to remove some words in the context but not others.
Specifically, while we remove adjectives which may carry signals of gender from the training corpora, we do not remove other parts of speech (e.g., verbs) under the reasoning that removing them may damage the training corpora too much for word2vec to effectively learn noun meanings.\footnote{Verbs may carry less signal for gender regardless. For example, \citealt{hoyle-etal-2019-unsupervised} find fewer significant differences in the usage of verbs than of adjectives towards people, and \citealt{williams-etal-2021-relationships} also report that verbs yielded smaller gender effects than adjectives.}
Future work can also explore improved representation methods for noun meaning.
For example, \citet{recski2016measuring} find that creating non-contextual word representations using a combination of word2vec, WordNet, and concept dictionaries can yield a better representation of meaning (i.e., achieving state-of-the-art correlation with the human-annotated similarity scores).
Third, the corpus choice (and subsequently the noun--adjective pairs on which we conduct our analysis) may factor into the results.
It is possible that when applied to other corpora (e.g., more colloquial ones like Reddit), this method may yield different results.
Fourth, the choice of languages analyzed further limits this study to languages with up to three gender classes. Future work can investigate languages with more complex gender systems.
Finally, our modeling approach assumes that the gender of a noun is influenced solely by its meaning. 
However, prior work has indicated that there are other factors that influence the grammatical gender of nouns such as their phonology and morphology \citep{corbett1991gender}.
Therefore, future work should investigate more complex graphical models in order to account for other confounding factors.

\section{Conclusion}

In this paper, we introduce a causal graphical model that jointly models the interactions between a noun's grammatical gender, its meaning, and adjective choice. 
We employ our model on five languages that exhibit grammatical gender to investigate the influence of nouns' gender on the adjectives chosen to describe those nouns. 
Replicating the findings of \citet{williams-etal-2021-relationships}, we find a substantial correlation between grammatical gender and adjective choice. 
However, taking advantage of our causal perspective, we show that when controlling for a noun's meaning, the effect of gender on adjective choice is marginal.
Thus, we provide further evidence against the neo-Whorfian hypothesis.\looseness=-1

\section*{Code Release}
The code and data necessary to replicate our empirical findings may be found at \url{https://github.com/rycolab/neo-whorf}.

\section*{Acknowledgments}
This work is mostly funded by Independent Research Fund Denmark under grant agreement number 9130-00092B. Thanks to Nadav Borenstein for proofreading an earlier manuscript draft.

\bibliography{anthology,custom}
\bibliographystyle{acl_natbib}

\appendix
\onecolumn

\section{Proof of Proposition 1}\label{sec:proof}

\jsmi*
\begin{mproof}
First, define the following distribution 
$m(\adj) \defeq \sum_{\gender \in \genders} \pgender(\gender) p(\adj \mid \mathrm{do}(\rvG=\gender))$.
Now, the result follows by algebraic manipulation
\begin{subequations}
\begin{align}
 \JSD_{\pgender}\Big(\Big\{p(\cdot &\mid \mathrm{do}(\rvG=\gender))\Big\}\Big) = \sum_{\gender \in \genders} \pgender(\gender)\, \KL\Big(p(\cdot \mid \mathrm{do}(\rvG=\gender)) \mid\mid m \Big)\\
    &= \sum_{\gender\in\genders} \pgender(\gender) \sum_{\adj \in \adjs} p(\adj \mid \mathrm{do}(\rvG = \gender)) \Big(\log p(\adj \mid \mathrm{do}(\rvG = \gender)) - \log m(\adj) \Big) \\
    &= \!
    \begin{multlined}[t][0.74\linewidth]
    \sum_{\gender\in\genders} \pgender(\gender) \sum_{\adj \in \adjs} p(\adj \mid \mathrm{do}(\rvG = \gender)) \log p(\adj \mid \mathrm{do}(\rvG = \gender)) - \\ \sum_{\gender\in\genders} \pgender(\gender) \sum_{\adj \in \adjs} p(\adj \mid \mathrm{do}(\rvG = \gender)) \log m(\adj)
    \end{multlined} \\
    %\end{align}
    %\begin{align}
    &= -\underbrace{\sum_{\gender\in\genders} \pgender(\gender) \mathrm{H}\left(\rvA \mid \mathrm{do}(\rvG = \gender\right))}_{\defeq \Hdo(\rvA \mid \rvG)} - \sum_{\gender\in\genders} \pgender(\gender) \sum_{\adj \in \adjs} p(\adj \mid \mathrm{do}(\rvG = \gender)) \log m(\adj)  \\
    &= -\Hdo\left(\rvA \mid \rvG \right) - \sum_{\gender\in\genders} \pgender(\gender) \sum_{\adj \in \adjs}  p(\adj \mid \mathrm{do}(\rvG = \gender))\log m(\adj)  \\
 &= -\Hdo\left(\rvA \mid \rvG \right) - \sum_{\adj \in \adjs} 
 % \underbrace{
 \sum_{\gender\in\genders}   \pgender(\gender)p(\adj \mid \mathrm{do}(\rvG = \gender))
 % }_{\defeq m(\adj)} 
 \log m(\adj)  \\
 &= -\Hdo\left(\rvA \mid \rvG \right) - \underbrace{\sum_{\adj \in \adjs} m(\adj) \log m(\adj)}_{\defeq -\Hdo(\rvA)}  \\
  &= -\Hdo\left(\rvA \mid \rvG \right) + \Hdo(\rvA)  = \Hdo(\rvA)  -\Hdo\left(\rvA \mid \rvG \right)  = \MIdo(\rvA; \rvG) 
\end{align}
\end{subequations}
\end{mproof}

\end{document}